\documentclass{llncs}
\usepackage[T1]{fontenc}
\usepackage{graphicx}
\usepackage{wrapfig}
\usepackage{amsmath}

\usepackage{xargs}   
\usepackage{xspace}
\usepackage[textsize=scriptsize,textwidth=25mm]{todonotes}
\usepackage{subcaption}
\usepackage{caption}
\usepackage{amssymb}

\definecolor{darkgrn}{rgb}{0, 0.75, 0}

\newcommand{\prelora}{\texttt{PreLoRA}}
\usepackage{algorithm}
\usepackage{algorithmic}

\begin{document}
\title{\texttt{PreLoRA}: Hybrid Pre-training of Vision Transformers with Full Training and Low-Rank Adapters}
\titlerunning{PreLoRA: Hybrid Pre-training of Vision Transformers}

\author{Krishu K Thapa$^{1}$, Reet Barik$^{2}$, Krishna Teja Chitty-Venkata$^{3}$, Murali Emani$^{2}$, Venkatram Vishwanath$^{2}$}
\authorrunning{Anonymous Authors}
\institute{Washington State University$^{1}$, Pullman, WA, USA \\
Argonne National Laboratory$^{2}$, Lemont, IL, USA \\
Red Hat$^{3}$, Raleigh, NC, USA}

%
%

%
\maketitle  
\pagestyle{plain}
%
\begin{abstract}
Training large models ranging from millions to billions of parameters 
is highly
resource-intensive, requiring significant time, compute, and memory.
It is observed that most of the learning (higher change in weights) takes place in the earlier stage of the training loop. As training progresses, these changes stabilize, suggesting that the resulting updates may be amenable to approximation using low intrinsic-rank matrices. Therefore, we propose an approach to identify such states of partial convergence and dynamically switch from full parameter training to Low Rank Adaptation (LoRA) on the \textit{ViT-Large} model. We introduce a flexible approach that leverages user-defined hyperparameters to determine the switching point and assign a rank specific to each module layer based on its level of convergence. Experimental results show that this approach preserves model accuracy while reducing the number of trainable parameters to 10\% of its original size, resulting in a 3$\times$ improvement in throughput, and a 1.5$\times$ reduction in average training time per epoch while also reducing GPU memory consumption by 20\%.
\end{abstract}

\section{Introduction}
\label{sec:intro}

Recent advances in machine learning have led to the development and deployment of large models across diverse domains \cite{lee2020biobert,radford2021learning,openai2023gpt4}. Their ever increasing size from millions to billions of parameters while enabling significant improvement in performance and generalization \cite{kaplan2020scaling}, has  introduced substantial computational and memory bottlenecks, rendering training prohibitively expensive \cite{rajbhandari,thompson2020computational}.

Parameter-efficient approaches have been explored to reduce trainable parameters while preserving model performance and generalization. Adapter methods \cite{houlsby2019parameter} insert small trainable modules into each layer with frozen base weights. However, this reduction comes at the cost of increased inference latency. Prefix and prompt tuning~\cite{li2021prefix,lester2021power} techniques optimize a small set of learnable task-specific vectors that are added to the input sequence or intermediate activations, but often exhibit limited expressivity when applied to complex tasks \cite{sun2024improving}. Low-Rank Adaptation (LoRA) \cite{Hu2021LoRALA} combines the flexibility of adapter methods with better expressivity than prompt-based methods, while maintaining minimal resource overhead by injecting low-rank trainable matrices into existing weight updates without adding any additional layers. While predominantly applied to fine-tuning, LoRA's integration into pretraining has the potential to accelerate training, reduce resource demands, and maintain model capacity.

\captionsetup{font={normalsize}}
\begin{figure*}[tb]
    \centering
    \begin{minipage}{0.45\textwidth} 
        \centering
        \includegraphics[scale=0.34]{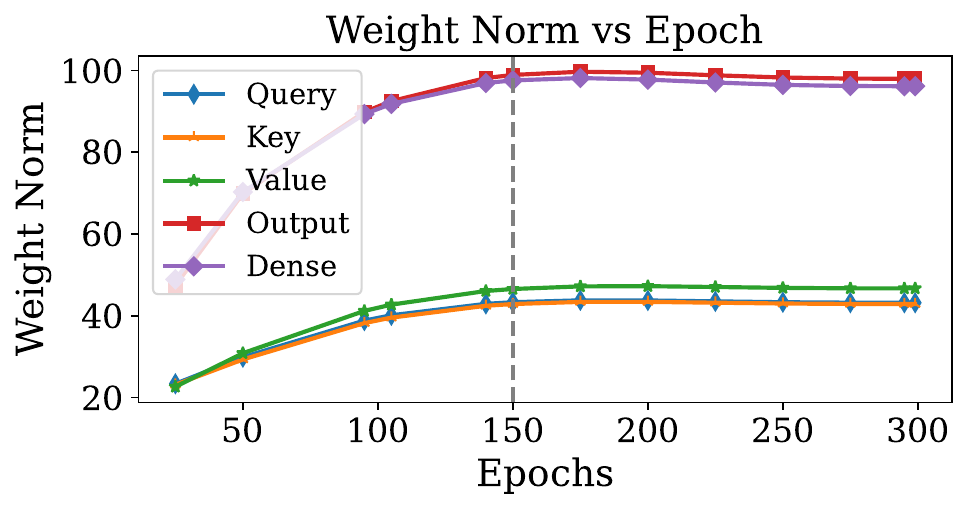}
        \caption*{(a)}
    \end{minipage}
    \hfill
    \begin{minipage}{0.45\textwidth}
        \centering
        \includegraphics[scale=0.34]{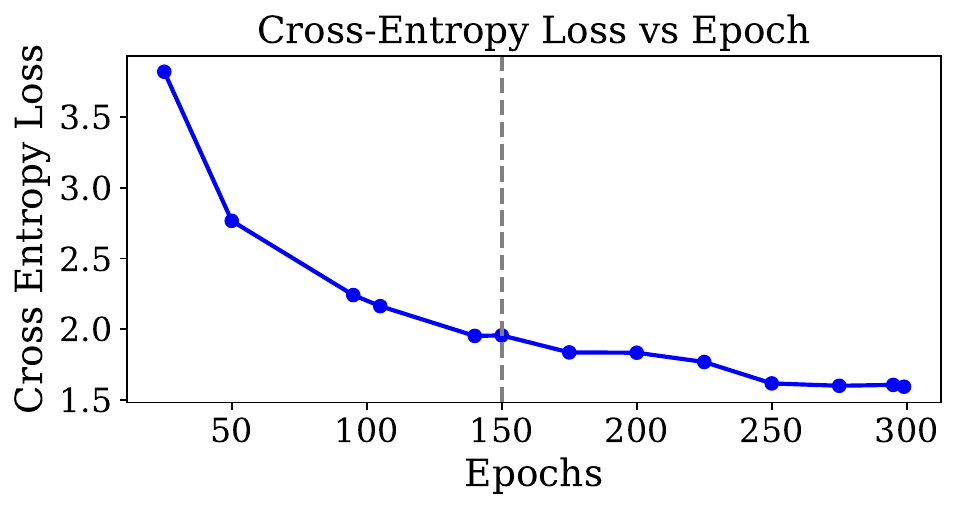} 
        \caption*{(b)}
    \end{minipage}    

    \caption{\small{\bf ViT-Large full model pretraining on ImageNet-1k}: (a) Weight norms of different modules over epochs. (b) Training Cross-Entropy Loss convergence over epochs. In both figures, the vertical dotted line highlights epoch 150 after which the weight norm change is minimal; however, the loss change is still significant.}
    \label{fig:motivation_fig}
    \vspace{-0.25in}
\end{figure*}

\noindent{\textbf{Motivation}:} During the initial training epochs, the model undergoes substantial weight updates, which gradually stabilize as training progresses. For a representative large Vision Transformer model,  Plot (a) of Figure~\ref{fig:motivation_fig} shows pronounced weight norm variations across all the modules initially followed by stabilization in the later epochs, while the model loss (shown in plot (b)) continues to decrease.
These observations suggest that, in later epochs, weight updates primarily reflect small refinements rather than large exploratory changes across all parameters. We hypothesize that the low-rank constraint of LoRA is sufficient to capture these refinements without significant accuracy loss.

\noindent{\textbf{Challenges}:} Identifying the optimal point to transition to LoRA-based training is non-trivial: an early transition can speed training but reduce accuracy, whereas a late transition preserves accuracy, but limits performance gains.  
A dynamic switching strategy, enabling explicit user control over this switching point, balances this accuracy–speed trade-off. Additionally, Figure~\ref{fig:motivation_fig} shows that while weight norms broadly exhibit consistent trends across modules, their magnitudes might differ considerably, indicating that a uniform low rank assignment may fail to capture module-specific convergence dynamics.

\noindent{\textbf{Contributions}:} In this work, we explore the integration of Low-Rank Adaptation (LoRA) into the pretraining of large-scale models. To that end, we have developed a framework, which we named \prelora{} that aims to address the challenges mentioned above. We use Vision Transformer (ViT-Large) \cite{Dosovitskiy2020AnII} with 300M parameters as a representative model for our experimental study. Our main contributions are as follows:
\begin{enumerate}
    \item We propose a custom convergence test (Section~\ref{subsec:part_conv_test}) that identifies the switching point from full-parameter to LoRA-based training using loss and weight-change thresholds and provides users explicit control over the accuracy–speed trade-off.

    \item We introduce a dynamic rank-assignment algorithm (Section~\ref{subsec:rank_assign}) that assigns module-specific ranks at the transition point, improving parameter efficiency.

    \item We conduct experiments on ViT-Large (Section~\ref{subsec:results}), to study how key design choices like the convergence criteria and warmup window size impact training efficiency and final model performance.
\end{enumerate}

These contributions form a generalizable framework for applying parameter-efficient techniques during pretraining, enabling faster and more resource-efficient training of large models across diverse domains. We further extend this framework by integrating an existing LoRA-based technique and present a targeted study demonstrating the complementary performance improvements it provides.

\section{Related Work}
\label{sec:rel_works}

Early work on adapters~\cite{he2022sparseadapter} established parameter-efficient fine-tuning (PEFT) by introducing lightweight, trainable modules into pretrained models, reducing the number of trainable parameters during adaptation. Numerous adapter variants have since been explored across NLP, computer vision, and speech recognition~\cite{he2022sparseadapter,chen2023hadamard,he2021effectiveness,pan2022st}. In parallel, prompt-tuning~\cite{lester2021power} and prefix-tuning~\cite{li2021prefix} emerged as architecture-preserving alternatives that append learnable prompts or prefixes to the input sequence, and have been widely adopted for fine-tuning large language and vision models under resource constraints~\cite{van2023open,choi2023codeprompt,vos2022towards,zhou2022learning,zhou2022conditional,jia2022visual}.
Our work instead centers on Low-Rank Adaptation (LoRA)~\cite{Hu2021LoRALA}, which provides a third paradigm by decomposing weight updates into low-rank matrices, thus achieving parameter efficiency without introducing new modules. LoRA has demonstrated superior efficiency, expressivity, and ease of integration compared to adapters and prompt-tuning~\cite{sun2024improving,dettmers2023qlora}, across multiple application domains~\cite{hayou2024lora+,lin2024tracking,wang2024lora,lu2023empirical,hamdi2021lora,dettmers2023qlora}.

Despite these advances, PEFT methods, including LoRA, have been studied primarily for fine-tuning, with 
only a small number of works having investigated the use of LoRA during pretraining from scratch. One work~\cite{10820024} applies LoRA to a spatiotemporal attention-based model on processing-in-memory (PIM) architectures by dynamically switching to a low-rank parameterization based on weight convergence. While this reduces training cost, it requires maintaining two model instances in parallel--a full model and its LoRA counterpart--and relies on statistical tests over their losses to detect convergence. This substantially increases memory usage and overhead, limiting scalability.
Another work, ReLoRA~\cite{lialin2023relora} provides an alternative way of incorporating low-rank updates during pre-training. After an initial warmup phase, the model transitions to a lower-rank parameterization and periodically merges and reinitializes low-rank parameters to emulate full-rank optimization. However, ReLoRA applies a fixed and pre-determined uniform rank across all layers and modules throughout the whole pretraining run. This approach does not account for variation in convergence dynamics across layers and modules, which flexible, layer- or module-specific rank assignments could better capture, potentially enabling more efficient parameter allocation and improved performance.

\section{PreLoRA}
\label{sec:method}

\begin{figure*}[tb]
    \centering
    \includegraphics[scale=0.43]{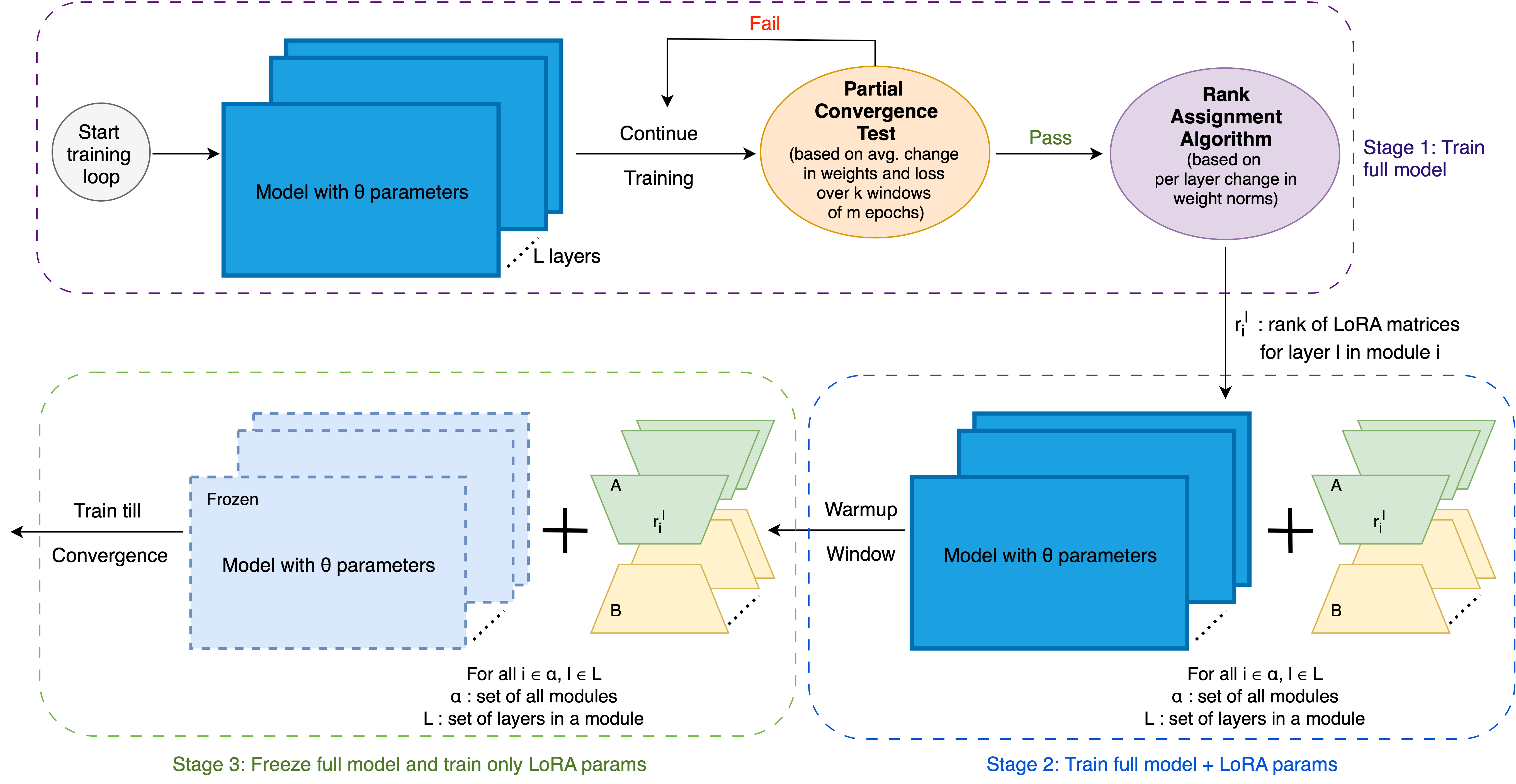}
    \caption{\small\prelora{} overall training workflow} 
    \label{fig:method}
\end{figure*}

Our approach relies on two key strategies. First, to determine an appropriate point in the training loop at which transitioning from full-parameter to LoRA-based training can be performed safely. To this end, we introduce a convergence criterion that assesses partial convergence. Additionally, to add flexibility, hyperparameters allow users to adjust the strictness of this criterion. Second, when switching to LoRA, we employ a rank-assignment algorithm that dynamically assigns ranks to individual layers of target modules based on the extent of weight norm convergence. Training is subsequently continued using the LoRA-augmented model. A schematic overview of \prelora's methodology can be found in Figure~\ref{fig:method}. Each component is described in detail in the subsections that follow.
\captionsetup{font={normalsize}}

\vspace{-0.3in}
\subsection{Partial Convergence Test}
\label{subsec:part_conv_test}

Determining an appropriate point in the training loop to switch from full-parameter to LoRA-based training is crucial. Premature switching may compromise model performance due to insufficient convergence of the modules, whereas delayed switching increases computation and prolongs training time. Therefore, we propose a partial convergence test, which provides a systematic mechanism to monitor module-level convergence during training and decide when the model has sufficiently converged and it is appropriate to transition to LoRA-based adaptation.

Algorithm~\ref{alg:conv_test} performs the convergence test at the module level, where $\alpha$ denotes the set of target modules for \prelora.  For $k$ windows of size $m$ epochs, $W_t^a$ for any module \(a \in \alpha\) and any window $(t \in k)$ denotes the weight norms across all its layers. Similarly, $L_t$ denotes the training loss for window $t$. The partial convergence test is considered passed if the percentage change in weight norms and losses between consecutive windows remain below $\tau$ and $\zeta$, respectively. $k$, $m$, $\tau$, and $\zeta$ are hyperparameters. The strictness of the test can be adjusted via hyperparameters: increasing $k$ and $m$ while decreasing $\tau$ and $\zeta$ makes the criterion more stringent, and vice versa.

\begin{algorithm}
\caption{Partial Convergence Test}
\label{alg:conv_test}
\begin{algorithmic}[1]
\STATE \textbf{Input:} Number of windows $k$, window size $m$, thresholds $\tau$, $\zeta$, module set $\alpha$. Weight norms $\{W_t^a\}$ for any module $a\in \alpha$ averaged across all layers, and losses $\{L_t\}$ at window $t \in k$.
\STATE \textbf{Output:}\textbf{True} if converged, \textbf{False} otherwise
\FOR{each module $a \in \alpha$}
    \FOR{$t = 2$ to $k$}
        \STATE Compute $\Delta W_t^a = \frac{\|W_t^a\| - \|W_{t-1}^a\|}{\|W_{t-1}^a\|} \times 100$
        \STATE Compute $\Delta L_t = \frac{L_t - L_{t-1}}{L_{t-1}} \times 100$
        \IF{$|\Delta W_t^a| > \tau$ or $|\Delta L_t| > \zeta$}
            \STATE \textbf{return} False
        \ENDIF
    \ENDFOR
\ENDFOR
\STATE \textbf{return} True
\end{algorithmic}
\end{algorithm}

\vspace{-0.3in}
\subsection{Rank Assignment Algorithm}
\label{subsec:rank_assign}


\begin{wrapfigure}{r}{0.49\textwidth}  
    \centering
    \vspace{-0.1in}
    \includegraphics[scale=0.23]{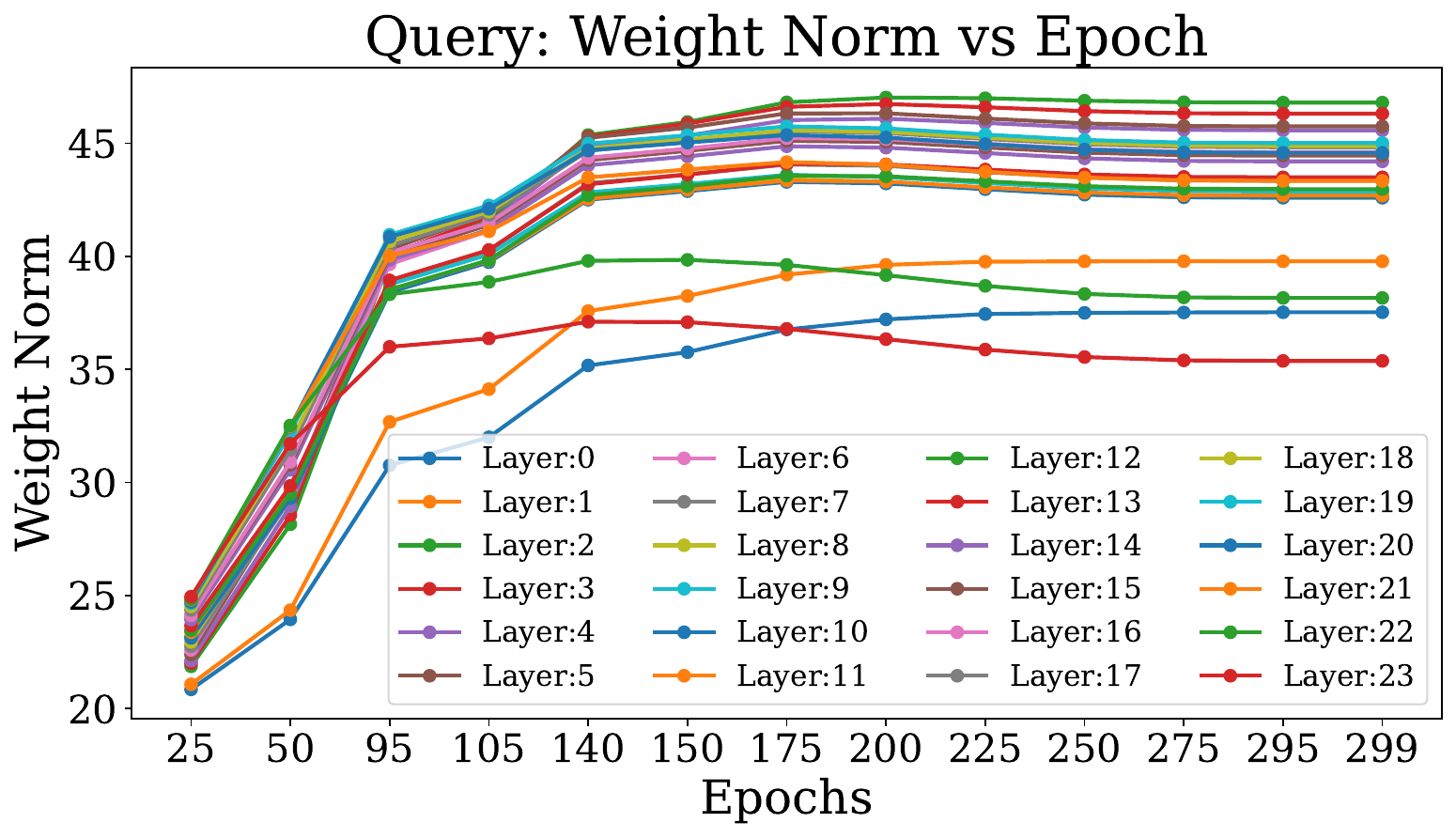}
    
    \caption{\small \textbf{ViT-Large}: weight norm of \textbf{\textit{Query}} module for all layers across training epochs.} 
    \label{fig:query_layer_norm}
    \vspace{-0.3in}
\end{wrapfigure}

The convergence test is performed at the module level, where 
\(\Delta W_t^a\) for any module \(a \in \alpha\) represents the average change; however, different layers may exhibit distinct rates of convergence, as seen in Figure~\ref{fig:query_layer_norm}. Therefore, ranks can be 
assigned to each layer based on its corresponding 
\(\Delta W_k^{a_l}\), where \(l \in L\) denotes the layer index, \(L\) the total number of layers in the model and $k$ the total number of windows considered in the partial convergence test. 

The rationale for the proposed dynamic ranking strategy is as follows:  following a successful partial convergence test over $k$ windows, the weights changes between the last two windows ($k-1$ and $k$), denoted by \(\Delta W_k^{a_l}\) can vary across different layers depending on their level of convergence. Therefore, layers that have substantially converged (smaller \(\Delta W_k^{a_l}\)) can be assigned lower ranks, whereas layers that are less converged (larger \(\Delta W_k^{a_l}\)) are allocated additional parameters through higher ranks. This approach ensures that more parameters are assigned to layers that still need substantial learning, while avoiding unnecessary parameters for already stable layers. 
The strategy used to determine the rank for different layers across all modules is described in Algorithm~\ref{alg:rank_assign}.

\begin{algorithm}
\caption{Rank Assignment Algorithm}
\label{alg:rank_assign}
\begin{algorithmic}[1]
\STATE \textbf{Input:} Minimum rank $r_{\min}$, maximum rank $r_{\max}$, weight norm changes $\Delta W_k^{a_l}$ for all modules $a \in \alpha$ and layers $l \in L$
\STATE \textbf{Output:} Layer-to-rank assignment function $\mathcal{A}: a_l \mapsto r$
\STATE Initialize rank set $\mathcal{R} \gets [\,]$
\FOR{$p = \log_2(r_{\min})$ to $\log_2(r_{\max})$}
    \STATE Append $2^p$ to $\mathcal{R}$
\ENDFOR
\STATE Initialize empty mapping $\mathcal{A} \gets \{\}$
\FOR{each module $a \in \alpha$}
    \STATE Collect weight norm changes for the module as, 
           $\textit{changes} \gets [\Delta W_k^{a_l} \;\; \forall l \in L]$
    \STATE min-max-norm($changes$) $\to$ $\mathcal{N}_a$ $\in$ [0,1] 
    \FOR{each layer $l \in L$ with normalized value $v \in \mathcal{N}_a$}
        \IF{$v \neq 0$}
            \STATE $i \gets \lceil v \cdot |\mathcal{R}| \rceil - 1$
        \ELSE
            \STATE $i \gets \lceil v \cdot |\mathcal{R}| \rceil$
        \ENDIF
        \STATE Assign layer $l$ of module $a$ to rank: 
               $\mathcal{A}[a_l] \gets \mathcal{R}[i]$
    \ENDFOR
\ENDFOR
\STATE \textbf{return} $\mathcal{A}$
\end{algorithmic}
\end{algorithm}
\vspace{-0.2in}

Ranks for each layer across modules are determined using a normalization-based bucketing strategy. Specifically, in line $9$ of Algorithm~\ref{alg:rank_assign}, for any module $a \in \alpha$, $changes$ contains all the weight changes $\Delta W_k^{a_l}$ for all layers $L$. This produces $\mathcal{N}_a$ (line $10$) by min-max-scaling the values in $changes$. For each layer $l \in L$ (loop on line $11$), the normalized weight change for each layer in $\mathcal{N}_a$ indexes into the array $\mathcal{R}$ containing the possible ranks (populated in line $4$-$5$ by enumerating all powers of $2$ between $r_{\min}$ and $r_{\max}$). The layer is subsequently assigned the rank at the corresponding index in $\mathcal{R}$ (line $17$). 

\subsection{Warmup (Full model + LoRA)}
The switch from the full parameter setting to the LoRA modules substantially reduces the number of trainable parameters. Since LoRA modules are randomly initialized, their updates may deviate from the expected optimization trajectory. This is particularly pronounced when employing LoRA while pre-training vs finetuning where the full parameter model has already converged. To mitigate this effect, the full model and LoRA modules are trained jointly for $w \in \mathbb{Z}$ warmup epochs, allowing LoRA modules to get guidance from the full model. Following the warmup period, the full model is frozen, and training proceeds only on the LoRA modules. The warmup window $w$ is a hyperparameter that regulates the influence of the full model on the initial learning of the LoRA modules.

\section{Experimental Results}
\label{sec:exp_res}

\subsection{Experimental Setup}
\label{sec:exp_setup}

\noindent{\textbf{Model and Data}:}To showcase the advantages of our \prelora{} framework, we employ the \textit{ViT-Large} architecture as a representative vision model, trained from scratch on the ImageNet-1k dataset. For an effective configuration to train ViT-Large from scratch on ImageNet-1K dataset, we adopt the training configuration prescribed by Steiner et al. \cite{steiner2021train}.

\noindent{\textbf{Hardware}:} Training was performed on a distributed supercomputing cluster using 64 NVIDIA A100 GPUs across 16 NVLink-connected nodes for high-bandwidth, low-latency communication. Each GPU provides 40 GB HBM2 memory with support for FP32, FP16, BF16, and INT8 precision. Compute nodes are equipped with 2.8 GHz AMD EPYC Milan 7543P (32-core) CPUs and 512 GB DDR4 RAM.

\noindent{\textbf{Software and Code}:} Experiments were implemented in PyTorch (v2.7.1) using HuggingFace Transformers (v0.33.0) with LoRA support via PEFT (v0.15.2). The open-source code, including data processing and visualization scripts, is available at \textcolor{blue}{\url{https://anonymous.4open.science/r/PreLoRA-D624}}.


\noindent{\textbf{Hyperparameters}:} Following Dahal et al.~\cite{10820024}, \prelora{} uses a partial convergence test with $k{=}3$ windows of $m{=}3$ epochs each. The target modules are $\alpha = {\text{[q (query), k (key), v (value), d (dense), o (output)]}}$. We study the impact of convergence thresholds $\tau$ and $\zeta$ on the accuracy–efficiency trade-off, and the warmup window $w$ for joint LoRA and full-model training before freezing base weights. Dynamic rank assignment is constrained between 8 and 64.

\subsection{Results}
\label{subsec:results}

\begin{figure}[t]
    \centering
    \begin{subfigure}[b]{0.48\linewidth}
        \centering
        \includegraphics[width=\linewidth]{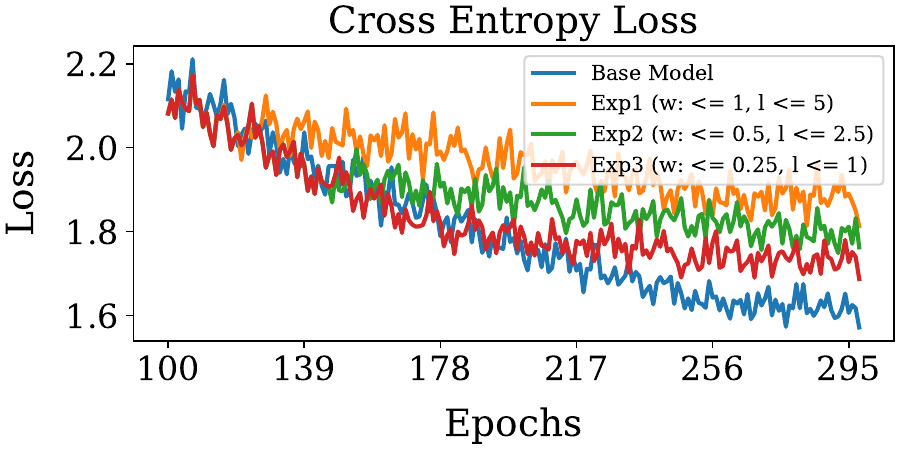}
        \caption{CE Loss}
        \label{fig:ce_loss}
    \end{subfigure}
    \hfill
    \begin{subfigure}[b]{0.48\linewidth}
        \centering
        \includegraphics[width=\linewidth]{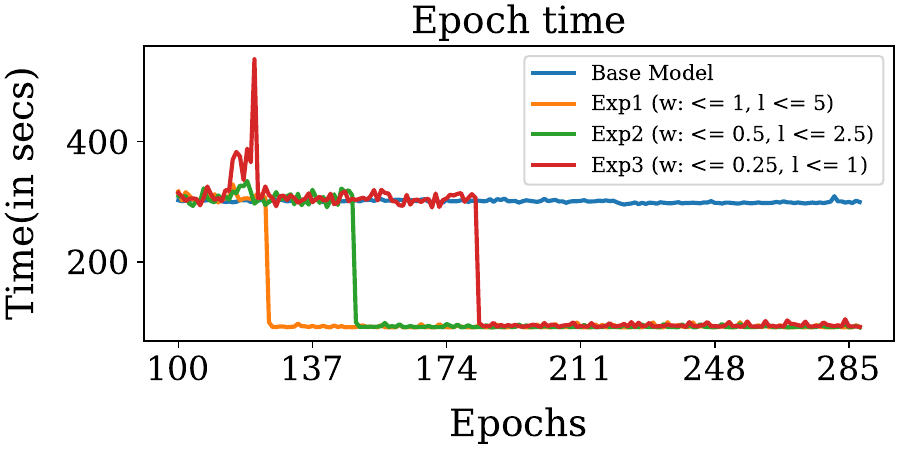}
        \caption{Training Time}
        \label{fig:train_time}
    \end{subfigure}

    \vspace{0.2em}

    \begin{subfigure}[b]{0.48\linewidth}
        \centering
        \includegraphics[width=\linewidth]{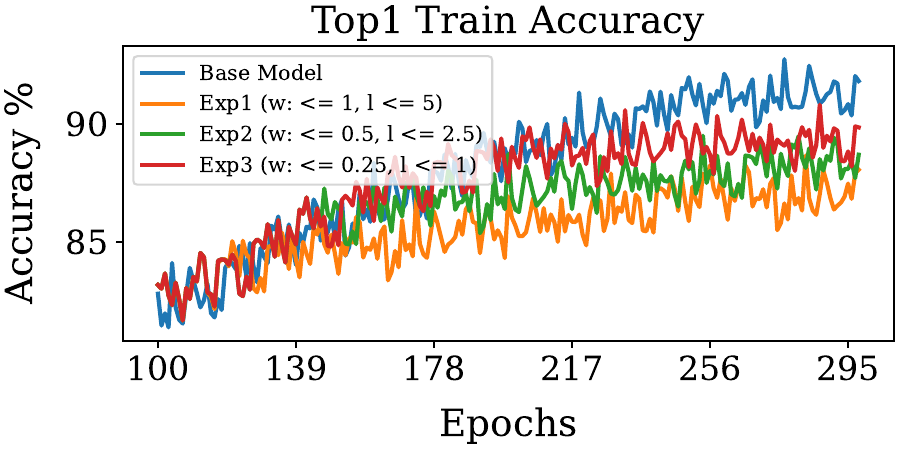}
        \caption{Top-1 Train Accuracy}
        \label{fig:top1_train}
    \end{subfigure}
    \hfill
    \begin{subfigure}[b]{0.48\linewidth}
        \centering
        \includegraphics[width=\linewidth]{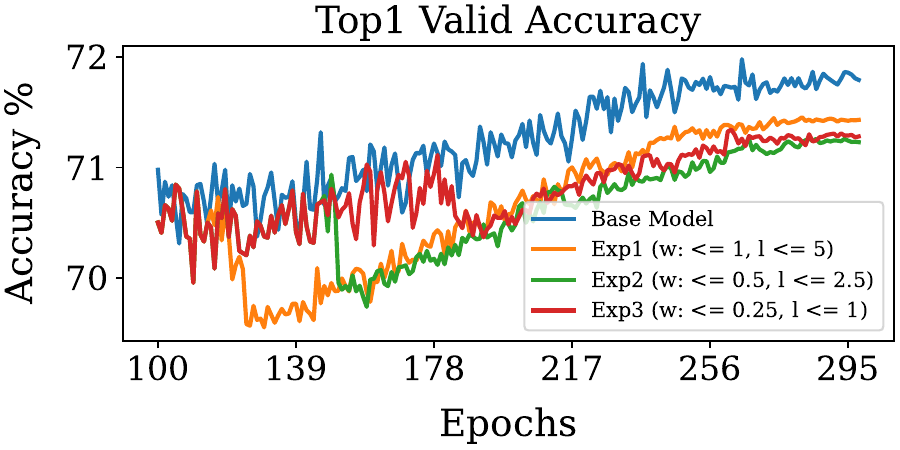}
        \caption{Top-1 Validation Accuracy}
        \label{fig:top1_valid}
    \end{subfigure}

    \caption{\small \textbf{Ablation Study.} Comparison of different \prelora{} settings to the full baseline model in terms of accuracy (a,c,d) and time (b). Displays epoch 100 and later for clarity}
    \label{fig:top1_accuracy}
    \vspace{-0.27in}
\end{figure}
We demonstrate the effectiveness of \prelora{} by comparing its performance across a range of hyperparameter settings against the full model baseline. To this end, this section
evaluates two key design choices: the strictness of the partial convergence test, determined by $\tau$ and $\zeta$, and the warmup window size $w$, which dictates how long the LoRA modules train alongside the full model before freezing its weights. 
This is followed by results that provide a broad overview of the performance gains achieved with \prelora{} in terms of time, throughput, and memory. Finally, we integrate ReLoRA’s merge-and-reinitialize technique to achieve further performance gains.
\vspace{-0.1in}

\subsubsection{Accuracy-Efficiency Trade-offs}
\label{subsec:ablation_study}
\leavevmode\par
\noindent \textbf{Strictness of partial convergence test ($\tau, \zeta$)}: As discussed previously in Section~\ref{subsec:part_conv_test}, determining an appropriate transition point from full-parameter training to LoRA-based training is a crucial component of the \prelora{} framework. In our method, this transition is guided by a convergence criterion defined through changes in weight norms ($\tau$) and training losses ($\zeta$). Together, these indicators capture partial convergence and trigger the switch.
We evaluate three partial convergence settings ranging from relaxed to strict thresholds: Exp1 $(\tau{=}1.0\%,, \zeta{=}5.0\%)$, Exp2 $(\tau{=}0.5\%,, \zeta{=}2.5\%)$, and Exp3 $(\tau{=}0.25\%,, \zeta{=}1.0\%)$.



A relaxed constraint causes the partial convergence criterion to be satisfied earlier, which is expected to yield substantial improvements in training efficiency, albeit at the cost of reduced accuracy, and vice versa. Figure~\ref{fig:train_time} shows that the most relaxed setup (Exp1) accelerates training by approximately 40\%, whereas the performance gain from the strictest setting (Exp3) is around 28\%. This gain, however, comes with a trade-off: the training cross-entropy loss (shown in Figure~\ref{fig:ce_loss}) converges to around 2.0 in Exp1, compared to 1.75 in Exp3. A similar trend is observed in training and validation accuracy (Figure~\ref{fig:top1_accuracy}).Despite the marginally degraded accuracy in terms of other metrics, Exp1 attains a higher top-1 validation accuracy than the other experiments. Therefore, for tasks where accuracy is paramount, switching later is advisable, whereas for applications that emphasize faster training and can tolerate a modest accuracy drop, an earlier switch is preferable. A significant observation is that, even during the pretraining phase, LoRA adaptation can capture the learning trend, despite a substantial reduction in the learnable parameters from 300M to 30M. 
\vspace{-0.1in}

\begin{figure}[thb]
    \centering
    \begin{subfigure}[b]{0.48\linewidth}  
        \centering
        \includegraphics[width=\linewidth]{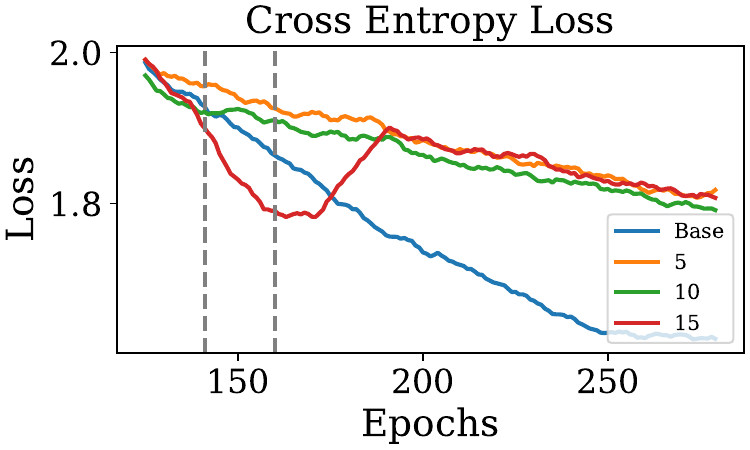}
        \caption{Cross-Entropy Loss}
        \label{fig:warmup_loss}
    \end{subfigure}
    \hfill
    \begin{subfigure}[b]{0.48\linewidth}  
        \centering
        \includegraphics[width=\linewidth]{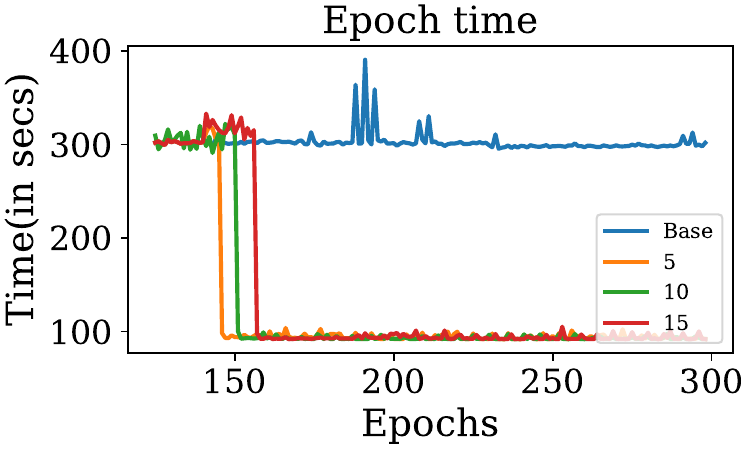}
        \caption{Epoch Speedup}
        \label{fig:warmup_speed}
    \end{subfigure}

    \caption{\small \textbf{Ablation Study.} Comparison of different LoRA warmup windows: (a) Cross-Entropy loss, and (b) training speedup.}
    \label{fig:warmup_ablation}
    \vspace{-0.2in}
\end{figure}
\subsubsection{Effect of warmup window size ($w$):} In this set of experiments, we vary the hyperparameter $w$ that controls the number of epochs (after passing the partial convergence test) we let the LoRA parameters train along with the full model before freezing the latter. The objective is to study the effect of the guidance provided by the full model to the LoRA parameters for different window sizes.
The set of values we use are $w = {5, 10, 15}$. We set $\tau = 0.50$ and $\zeta = 2.5$ corresponding to Exp2. This setting results in the training mode switching to only the LoRA parameters (with the base model frozen) at around the $150$-th epoch. We know from Figure~\ref{fig:motivation_fig} that it is during this phase of the training cycle that the base model weights largely stabilize for it to be able to provide the necessary guidance. The results are shown in Figure~\ref{fig:warmup_ablation}. 
We present the loss curves (Figure~\ref{fig:warmup_loss}) and epoch times (Figure~\ref{fig:warmup_speed}) for warmup window sizes $w \in {5, 10, 15}$, with the full model serving as the baseline, evaluated from epoch 135 onward. The epoch times exhibit the expected trend: a shorter warmup window triggers an earlier transition to LoRA-only training, yielding greater performance gains.


    

\begin{wrapfigure}{r}{0.54\textwidth}  
    \centering
    \vspace{-0.20in}
    \begin{subfigure}[b]{0.48\linewidth}
        \centering
        \includegraphics[width=\linewidth]{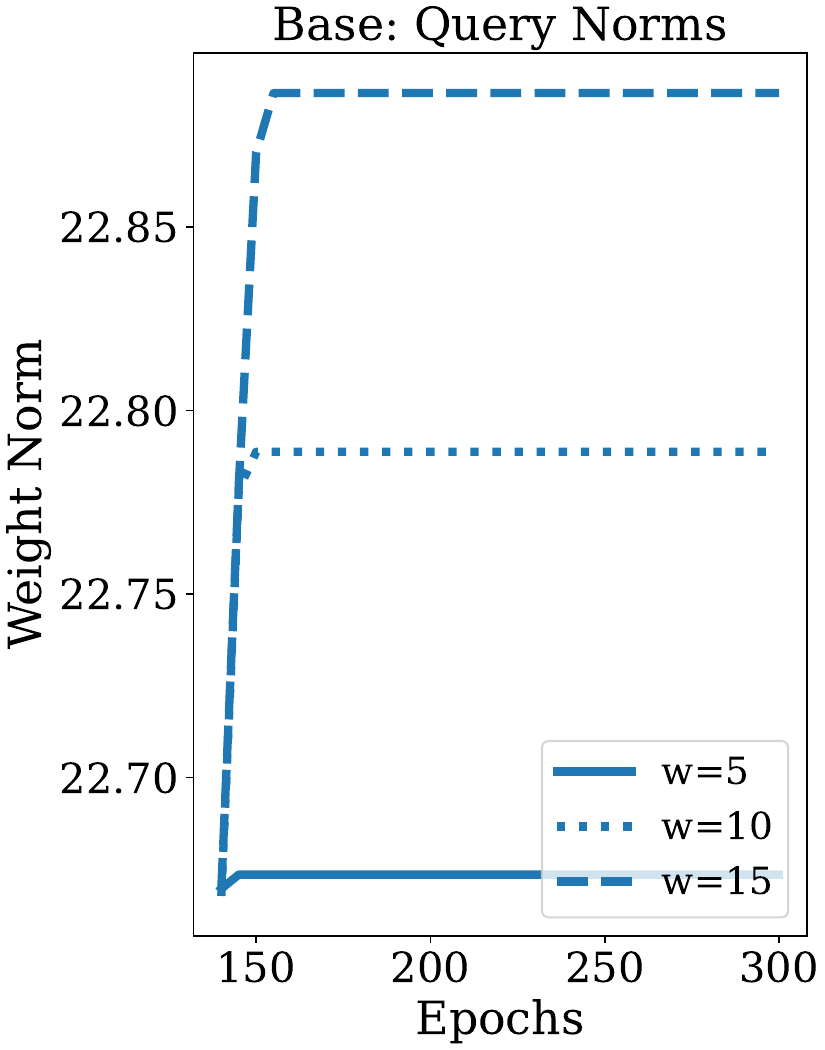}
        \caption{}
        \label{fig:query_base_norm}
    \end{subfigure}
    \hfill
    \begin{subfigure}[b]{0.48\linewidth}
        \centering
        \includegraphics[width=\linewidth]{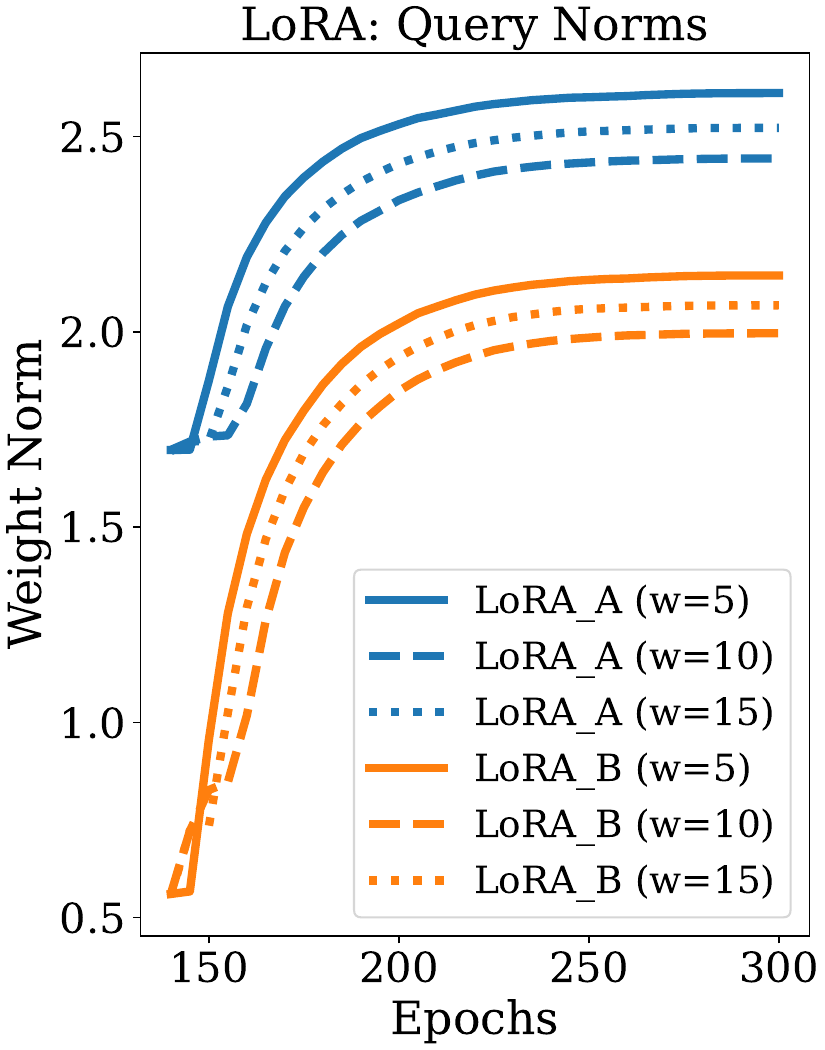}
        \caption{}
        \label{fig:query_lora_norm}
    \end{subfigure}
    
    \caption{\small Effect of 
    Weight norms (Query) on warmup windows for (a) the Base model 
    and (b) LoRA parameters.}
    \label{fig:warmup_norms}
    \vspace{-0.3in}
\end{wrapfigure}
For cross-entropy loss, the baseline slightly outperforms the \prelora{} variants overall. The w=15 configuration briefly surpasses the baseline during the warmup phase (Figure~\ref{fig:warmup_loss}) before converging with other settings. Short warmup windows may insufficiently guide LoRA adaptation, whereas longer warmup periods shift most updates to the base model, limiting LoRA learning. Hence, w=15 shows a temporary post-warmup loss increase before stabilizing and converging.


To examine this, we analyzed weight norm dynamics of the base and LoRA parameters. Figure~\ref{fig:query_base_norm} shows that a larger warmup window (w=15) increases base model weight norms, while Figure~\ref{fig:query_lora_norm} shows that LoRA norms decrease with longer windows. Overall, longer warmups offer limited accuracy gains and slower improvements, with w=10 providing the best trade-off between efficiency and final performance.

\subsubsection{Time, compute, and memory:}
\label{subsec:tcm}

        

\begin{wrapfigure}{r}{0.5\textwidth}  
    \centering
    \includegraphics[scale=0.35]{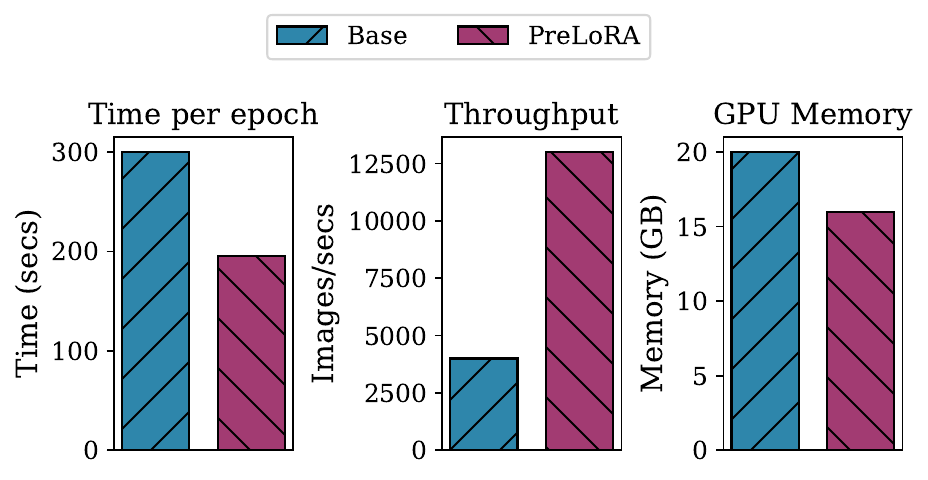}
    \caption{\small \textbf{Full model (Base) vs \prelora{}:} 
    time, compute, and memory utilization}
    \label{fig:bar_charts_row}
    \vspace{-0.2in}
\end{wrapfigure}

LoRA adaptation reduces the number of trainable parameters, making backpropagation more time- and memory-efficient. In our setup, trainable parameters decrease from 300M to about 10\% of the full model. Consequently, \prelora{} improves training efficiency in time, computation, and memory usage. As shown in Figure~\ref{fig:bar_charts_row}, \prelora{} achieves a 1.5$\times$ reduction in average epoch time, resulting in approximately 9 hours of savings over 300 epochs. This leads to a 3$\times$ increase in throughput and a 20\% reduction in GPU memory usage, enabling the potential use of larger batch sizes.
\vspace{-0.1in}

\subsubsection{Integration with ReLoRA:}
\label{subsec:relora}




\begin{wrapfigure}{r}{0.40\textwidth}  
    \centering
    \includegraphics[width=\linewidth,
                     height=3.5cm,
                     keepaspectratio]{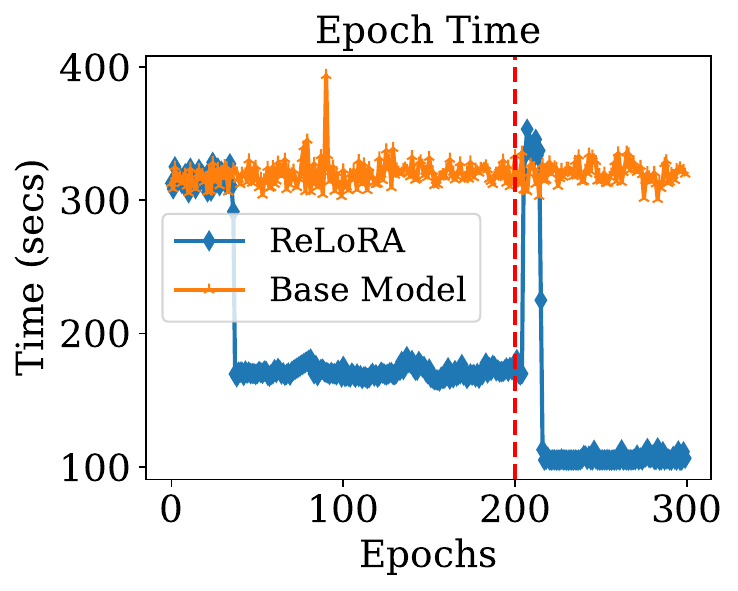}
    \caption{\small \textbf{ReLoRA Integration}: Time Per Epoch }
    \label{fig:relora_et}
    \vspace{-0.3in}
\end{wrapfigure}
\prelora{} reduces parameter count after partial convergence, typically in the later training stages. To explore whether earlier epochs can also benefit, we integrate the ReLoRA strategy~\cite{lialin2023relora} from the start, beginning with several full-parameter epochs before transitioning to periodic merge-and-reinitialize updates. 
After the first transition, the rank is reduced from 1024 to 256 (300M to 114M parameters), continuing until partial convergence. At this point our rank assignment algorithm sets layer-specific rank between 8 and 64, further reducing the parameter count to less than 10\% of the original with training continuing at these ranks for the remainder of the run. This hybrid approach preserves validation accuracy with minimal impact on cross-entropy loss, improves training speed by $\sim$ 122 seconds per epoch (42\% performance gain), as shown in Figure~\ref{fig:relora_et}. All experiments use the Exp1 hyperparameters with a 10-epoch warmup following convergence detection. Epoch times show a temporary rise from epochs 200 to 210, corresponding to the 10-epoch warmup after partial convergence.

\section{Conclusion}
\label{sec:conclusion}

In summary, we present and evaluate a LoRA-based approach for reducing trainable parameters during large vision model pre-training. Our results show that, under appropriate convergence criteria, low-rank constraints are sufficient to capture late-stage training refinements without significant accuracy loss. The method identifies an effective transition strategy between full-parameter and low-parameter training, giving users explicit control over the accuracy–speed trade-off. Although evaluated on ViT-Large, the approach generalizes to other architectures and is further enhanced through integration with ReLoRA. Future work will explore online hyperparameter optimization to establish a principled convergence criterion and improve robustness and scalability across domains.

\section{Acknowledgment}
\label{sec:ack}

This research used resources of the Argonne Leadership Computing Facility, a U.S. Department of Energy (DOE) Office
of Science user facility at Argonne National Laboratory and
is based on research supported by the U.S. DOE Office of
Science-Advanced Scientific Computing Research Program,
under Contract No. DE-AC02-06CH11357

{
\small
\bibliographystyle{splncs04}
\bibliography{main}

@article{lee2020biobert,
  title={BioBERT: a pre-trained biomedical language representation model for biomedical text mining},
  author={Jinhyuk Lee and others},
  journal={Bioinformatics},
  volume={36},
  number={4},
  pages={1234--1240},
  year={2020},
  doi={10.1093/bioinformatics/btz682},
  url={https://academic.oup.com/bioinformatics/article/36/4/1234/5566506}
}

@article{radford2021learning,
  title={Learning Transferable Visual Models From Natural Language Supervision},
  author={Alec Radford and others},
  journal={Proceedings of the International Conference on Machine Learning (ICML)},
  year={2021},
  url={https://openai.com/research/clip}
}

@article{openai2023gpt4,
  author = {OpenAI},
  title = {GPT-4 Technical Report},
  year = {2023},
  journal = {arXiv preprint arXiv:2303.08774},
  url = {https://arxiv.org/abs/2303.08774}
}

@article{kaplan2020scaling,
  title={Scaling Laws for Neural Language Models},
  author={Kaplan, Jared and McCandlish, Sam and Henighan, Tom and Brown, Tom B and Chess, Benjamin and Child, Rewon and Gray, Scott and Radford, Alec and Wu, Jeffrey and Amodei, Dario},
  journal={arXiv preprint arXiv:2001.08361},
  year={2020},
  url={https://arxiv.org/abs/2001.08361}
}

@INPROCEEDINGS{rajbhandari,
  author={Rajbhandari, Samyam and Rasley, Jeff and Ruwase, Olatunji and He, Yuxiong},
  booktitle={SC20: International Conference for High Performance Computing, Networking, Storage and Analysis}, 
  title={ZeRO: Memory optimizations Toward Training Trillion Parameter Models}, 
  year={2020},
  volume={},
  number={},
  pages={1-16},
  doi={10.1109/SC41405.2020.00024}}

@article{thompson2020computational,
  title={The computational limits of deep learning},
  author={Thompson, Neil C and Greenewald, Kristjan and Lee, Keeheon and Manso, Gabriel F},
  journal={arXiv preprint arXiv:2007.05558},
  year={2020}
}

@inproceedings{houlsby2019parameter,
  title={Parameter-efficient transfer learning for NLP},
  author={Houlsby, Neil and Giurgiu, Andrei and Stojnic, Rami and Hu, Jian and Tsai, Yi and Kim, Danqi and Grefenstette, Edward and Ruder, Sebastian and Stojanovic, Jure and Vuli{\'c}, Ivan and others},
  booktitle={Proceedings of the 36th International Conference on Machine Learning},
  pages={2790--2799},
  year={2019}
}

@inproceedings{li2021prefix,
  title={Prefix-Tuning: Optimizing Continuous Prompts for Generation},
  author={Li, Xiang and Liang, Percy},
  booktitle={Proceedings of the 38th International Conference on Machine Learning},
  pages={5390--5400},
  year={2021}
}

@article{Hu2021LoRALA,
  title={LoRA: Low-Rank Adaptation of Large Language Models},
  author={J. Edward Hu and Yelong Shen and Phillip Wallis and Zeyuan Allen-Zhu and Yuanzhi Li and Shean Wang and Weizhu Chen},
  journal={ArXiv},
  year={2021},
  volume={abs/2106.09685},
  url={https://api.semanticscholar.org/CorpusID:235458009}
}

@article{Dosovitskiy2020AnII,
  title={An Image is Worth 16x16 Words: Transformers for Image Recognition at Scale},
  author={Alexey Dosovitskiy and Lucas Beyer and Alexander Kolesnikov and Dirk Weissenborn and Xiaohua Zhai and Thomas Unterthiner and Mostafa Dehghani and Matthias Minderer and Georg Heigold and Sylvain Gelly and Jakob Uszkoreit and Neil Houlsby},
  journal={ArXiv},
  year={2020},
  volume={abs/2010.11929},
  url={https://api.semanticscholar.org/CorpusID:225039882}
}

@article{he2022sparseadapter,
  title={Sparseadapter: An easy approach for improving the parameter-efficiency of adapters},
  author={He, Shwai and Ding, Liang and Dong, Daize and Zhang, Miao and Tao, Dacheng},
  journal={arXiv preprint arXiv:2210.04284},
  year={2022}
}

@inproceedings{chen2023hadamard,
  title={Hadamard adapter: An extreme parameter-efficient adapter tuning method for pre-trained language models},
  author={Chen, Yuyan and Fu, Qiang and Fan, Ge and Du, Lun and Lou, Jian-Guang and Han, Shi and Zhang, Dongmei and Li, Zhixu and Xiao, Yanghua},
  booktitle={Proceedings of the 32nd ACM international conference on information and knowledge management},
  pages={276--285},
  year={2023}
}

@article{he2021effectiveness,
  title={On the effectiveness of adapter-based tuning for pretrained language model adaptation},
  author={He, Ruidan and Liu, Linlin and Ye, Hai and Tan, Qingyu and Ding, Bosheng and Cheng, Liying and Low, Jia-Wei and Bing, Lidong and Si, Luo},
  journal={arXiv preprint arXiv:2106.03164},
  year={2021}
}

@article{pan2022st,
  title={St-adapter: Parameter-efficient image-to-video transfer learning},
  author={Pan, Junting and Lin, Ziyi and Zhu, Xiatian and Shao, Jing and Li, Hongsheng},
  journal={Advances in Neural Information Processing Systems},
  volume={35},
  pages={26462--26477},
  year={2022}
}

@article{lester2021power,
  title={The power of scale for parameter-efficient prompt tuning},
  author={Lester, Brian and Al-Rfou, Rami and Constant, Noah},
  journal={arXiv preprint arXiv:2104.08691},
  year={2021}
}

@inproceedings{van2023open,
  title={Open-ended medical visual question answering through prefix tuning of language models},
  author={Van Sonsbeek, Tom and Derakhshani, Mohammad Mahdi and Najdenkoska, Ivona and Snoek, Cees GM and Worring, Marcel},
  booktitle={International Conference on Medical Image Computing and Computer-Assisted Intervention},
  pages={726--736},
  year={2023},
  organization={Springer}
}

@inproceedings{choi2023codeprompt,
  title={CodePrompt: Task-agnostic prefix tuning for program and language generation},
  author={Choi, YunSeok and Lee, Jee-Hyong},
  booktitle={Findings of the Association for Computational Linguistics: ACL 2023},
  pages={5282--5297},
  year={2023}
}

@article{dettmers2023qlora,
  title={Qlora: Efficient finetuning of quantized llms},
  author={Dettmers, Tim and Pagnoni, Artidoro and Holtzman, Ari and Zettlemoyer, Luke},
  journal={Advances in neural information processing systems},
  volume={36},
  pages={10088--10115},
  year={2023}
}

@inproceedings{vos2022towards,
  title={Towards parameter-efficient automation of data wrangling tasks with prefix-tuning},
  author={Vos, David and D{\"o}hmen, Till and Schelter, Sebastian},
  booktitle={NeurIPS 2022 First Table Representation Workshop},
  pages={1--9},
  year={2022}
}

@article{zhou2022learning,
  title={Learning to prompt for vision-language models},
  author={Zhou, Kaiyang and Yang, Jingkang and Loy, Chen Change and Liu, Ziwei},
  journal={International Journal of Computer Vision},
  volume={130},
  number={9},
  pages={2337--2348},
  year={2022},
  publisher={Springer}
}

@inproceedings{zhou2022conditional,
  title={Conditional prompt learning for vision-language models},
  author={Zhou, Kaiyang and Yang, Jingkang and Loy, Chen Change and Liu, Ziwei},
  booktitle={Proceedings of the IEEE/CVF conference on computer vision and pattern recognition},
  pages={16816--16825},
  year={2022}
}

@inproceedings{jia2022visual,
  title={Visual prompt tuning},
  author={Jia, Menglin and Tang, Luming and Chen, Bor-Chun and Cardie, Claire and Belongie, Serge and Hariharan, Bharath and Lim, Ser-Nam},
  booktitle={European conference on computer vision},
  pages={709--727},
  year={2022},
  organization={Springer}
}

@article{sun2024improving,
  title={Improving lora in privacy-preserving federated learning},
  author={Sun, Youbang and Li, Zitao and Li, Yaliang and Ding, Bolin},
  journal={arXiv preprint arXiv:2403.12313},
  year={2024}
}

@article{hayou2024lora+,
  title={Lora+: Efficient low rank adaptation of large models},
  author={Hayou, Soufiane and Ghosh, Nikhil and Yu, Bin},
  journal={arXiv preprint arXiv:2402.12354},
  year={2024}
}

@article{wang2024lora,
  title={Lora-ga: Low-rank adaptation with gradient approximation},
  author={Wang, Shaowen and Yu, Linxi and Li, Jian},
  journal={Advances in Neural Information Processing Systems},
  volume={37},
  pages={54905--54931},
  year={2024}
}

@inproceedings{lin2024tracking,
  title={Tracking meets lora: Faster training, larger model, stronger performance},
  author={Lin, Liting and Fan, Heng and Zhang, Zhipeng and Wang, Yaowei and Xu, Yong and Ling, Haibin},
  booktitle={European Conference on Computer Vision},
  pages={300--318},
  year={2024},
  organization={Springer}
}

@article{lu2023empirical,
  title={An empirical study of scaling instruct-tuned large multimodal models},
  author={Lu, Yadong and Li, Chunyuan and Liu, Haotian and Yang, Jianwei and Gao, Jianfeng and Shen, Yelong},
  journal={arXiv preprint arXiv:2309.09958},
  year={2023}
}

@article{hamdi2021lora,
  title={LoRa-RL: Deep reinforcement learning for resource management in hybrid energy LoRa wireless networks},
  author={Hamdi, Rami and Baccour, Emna and Erbad, Aiman and Qaraqe, Marwa and Hamdi, Mounir},
  journal={IEEE Internet of Things Journal},
  volume={9},
  number={9},
  pages={6458--6476},
  year={2021},
  publisher={IEEE}
}

@ARTICLE{10820024,
  author={Dahal, Saiman and Dhingra, Pratyush and Thapa, Krishu Kumar and Pande, Partha Pratim and Kalyanaraman, Ananth},
  journal={IEEE Transactions on Parallel and Distributed Systems}, 
  title={HpT: Hybrid Acceleration of Spatio-Temporal Attention Model Training on Heterogeneous Manycore Architectures}, 
  year={2025},
  volume={36},
  number={3},
  pages={407-421},
  doi={10.1109/TPDS.2024.3522781}}

@article{steiner2021train,
  title={How to train your vit? data, augmentation, and regularization in vision transformers},
  author={Steiner, Andreas and Kolesnikov, Alexander and Zhai, Xiaohua and Wightman, Ross and Uszkoreit, Jakob and Beyer, Lucas},
  journal={arXiv preprint arXiv:2106.10270},
  year={2021}
}

@article{lialin2023relora,
  title={Relora: High-rank training through low-rank updates},
  author={Lialin, Vladislav and Shivagunde, Namrata and Muckatira, Sherin and Rumshisky, Anna},
  journal={arXiv preprint arXiv:2307.05695},
  year={2023}
}
}

\end{document}